# Precipitation Nowcasting: Leveraging bidirectional LSTM and 1D CNN

Maitreya Patel [*], Anery Patel [+], Dr. Ranendu Ghosh [×]
*Dhirubhai Ambani Institute of Information and Communication Technology*

*Abstract*— Short-term rainfall forecasting, also known as precipitation nowcasting has become a potentially fundamental technology impacting significant real-world applications ranging from flight safety, rainstorm alerts to farm irrigation timings. Since weather forecasting involves identifying the underlying structure in a huge amount of data, deep-learning based precipitation nowcasting has intuitively outperformed the traditional linear extrapolation methods. Our research work intends to utilize the recent advances in deep learning to nowcasting, a multi-variable time series forecasting problem. Specifically, we leverage a bidirectional LSTM (Long Short-Term Memory) neural network architecture which remarkably captures the temporal features and long-term dependencies from historical data. To further our studies, we compare the bidirectional LSTM network with 1D CNN model to prove the capabilities of sequence models over feed-forward neural architectures in forecasting related problems.

I. INTRODUCTION

Originally defined by Keith Browning, nowcasting is "the description of the current state of the weather in detail and the prediction of changes that can be expected on a timescale of a few hours". Precipitation Nowcasting refers to the problem of analyzing and forecasting regional rainfall for a very short-term (eg., 0-6 hours into future) with spatial accuracy on the order of 1 kilometer. It reduces the uncertainty of the day-ahead forecast of the previous day and plays a crucial role in risk prevention and crisis management. Disasters at small scales and short time periods like hail storms and flash floods, along with sub-events within large-scaled catastrophes can be detected and well managed. Among other possibilities, it can help farmers to avoid lately sprayed fertilizers getting washed away, facilitate drivers to pre-determine road conditions and can give weather guidance for flight safety in regional aviation.

It demands a heuristic approach which cannot be solved by typical Numerical Weather Prediction (NWP) problems. Due to the dynamic and non-linear nature of nowcasting, linear extrapolation techniques cannot capture the underlying patterns of aberrations and trends based on the historical data. Nowcasting being a heavy data-oriented field of study, we can exploit the processing capabilities of sequence-based deep neural networks.

In the standard terms of machine learning, this type of problem is studied under the domain of Time Series Analysis. Time series is a form of data taken sequentially over time. In order to apply supervised machine learning models, we need to extract features from a given time series. In this work, we first transformed the input time series data into a set of features. A set of temporal features used for predictions are obtained by training a bidirectional LSTM on the processed data. Subsequently, the extracted temporal features are used to train an artificial neural network (ANN) based classifier to predict the possibility of rainfall in a specific geographic region[10]. The accuracy of the predictions confirms the promising performance of sequence-based deep learning algorithms.

Our second approach includes a specialized convolutional network i.e. 1-D CNN for temporal feature extraction and classification similar to our first approach. It has been observed that the former approach takes slightly lesser time for convergence and has given better accuracy with a smoother loss function. The approach proposed is restricted to a particular geographical area and can be further adapted.

Section II describes the datasets used for measuring the performance of the architectures. Section III deals with the details of the neural network architecture of bidirectional LSTM and 1D CNN. It is subsequently followed by the details of our experiments, results and conclusions.

[1]*patel.maitreya57@gmail.com
+anery.daiict@gmail.com
x ranendu_ghosh@daiict.ac.in

## II. DATA PIPELINE

### A. DATA SOURCE

The data used in this study includes the following:
1. Indian weather pattern: The hourly time series data used to train the networks is available for Ahmedabad airport specifically, from 2010 to 2017. Given that the monsoon in India typically extends from late July till early September, the data used for training includes only the aforementioned four months. The weather condition parameters in the data include temperature, wind speed, pressure, and humidity.[1]
2. North-America weather pattern: In order to check the cross-dataset performance and generic nature of our models, we considered using another hourly time series data publicly available on Kaggle under the name "Historical Hourly Weather Data 2012-2017".[2] It includes multiple weather condition parameters like humidity, pressure, temperature, wind speed and wind direction for different cities. However, to precisely gauge the predictions, data for only those cities have been considered where rainfall occurs more often and in measurable amount than cities like Vancouver, Seattle and Portland.

### B. DATA PREPROCESSING

Time Series data must be re-framed as a supervised learning dataset before we can start using machine learning algorithms. There is no concept of input and output features in time series. Instead, we must choose the variable to be predicted and use feature engineering to construct all of the inputs that will be used to make predictions for future time steps.

Table 2 shows the sample raw weather dataset collected for Ahmedabad region. Table 1 shows the processed dataset containing distinct features suitable for training a supervised learning model. Suppose at present time $(t)$, we need to determine the weather conditions for the next hour $(t+1)$. We consider all the 5 parameters (temperature, wind speed, humidity, pressure, and rainfall) given for past 24 hours. The input time series has 5 parametric values for each timestep (an hour). We transform it into five features each having 24 values. Each row now represents a different day. The total input parameters thus, sums to $5 * 24 = 120$. As shown in Table 1, this transformed dataset will convert our problem into a regression problem. The input variables are $var1$, $var2$, $var3$, $var4$ and $var5$ with each variable having 24 values for each hour of the day. The $var5(t+1)$ denotes the predicted value of rainfall for the next hour $(t+1)$ using the past input data.

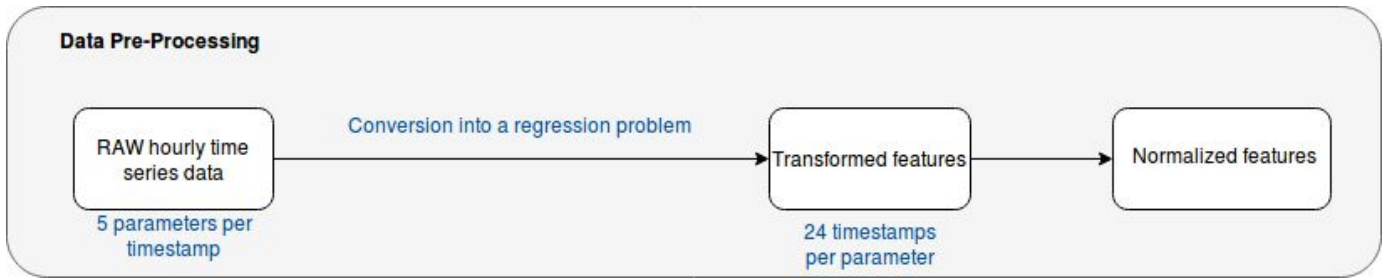

**Figure A. Flow chart on data preprocessing**

| Var1 (t-24) | Var2 (t-24) | Var3 (t-24) | Var4 (t-24) | Var5 (t-24) | ... | Var1 (t) | Var2 (t) | Var3 (t) | Var4 (t) | Var5 (t) | Var5 (t+1) |
|---|---|---|---|---|---|---|---|---|---|---|---|
| 37 | 19 | 42 | 1000 | 0 | ... | 27 | 13 | 84 | 1004 | 1 | 0 |
| 37 | 19 | 35 | 1000 | 0 | ... | 26 | 19 | 74 | 1005 | 0 | 0 |
| 39 | 24 | 31 | 998 | 0 | ... | 26 | 17 | 79 | 1005 | 0 | 1 |

**Table 1. Sample Transformed features**

| Year | Month | Date | Time | Temp(°C) | Wind speed (km/h) | Humidity (%) | Pressure (mbar) | Rainfall (1/0) |
|------|-------|------|------|----------|-------------------|--------------|-----------------|----------------|
| 2010 | 7 | 15 | 08:10 | 27 | 13 | 84 | 1004 | 1 |
| 2010 | 7 | 15 | 08:40 | 26 | 19 | 74 | 1005 | 0 |
| 2010 | 7 | 15 | 09:10 | 26 | 17 | 79 | 1005 | 0 |
| 2010 | 7 | 15 | 09:40 | 26 | 13 | 84 | 1005 | 1 |
| 2010 | 7 | 15 | 10:10 | 26 | 11 | 89 | 1004 | 1 |
| 2010 | 7 | 15 | 10:40 | 25 | 13 | 86 | 1004 | 0 |

**Table 2. Sample raw hourly time series**

### III. MODEL

To tackle the problem of rainfall nowcasting, we have developed two models which takes past 12-24 hours time series data of one station and predicts weather conditions for 1-3 hours ahead into future.

#### A. BIDIRECTIONAL LSTM

LSTM based architecture are very popular for their efficiency on temporal feature extraction from any type of time series data. Sequence based models like recurrent neural networks and LSTMs take as their input not just the current input example they see, but also what they have perceived previously in time. The model preserves information from inputs that has already passed through it using the hidden state. LSTMs have an edge over conventional feed-forward neural networks and Recurrent neural Networks (RNN)[7] because of their property of selectively remembering patterns for long durations of time. [3]

| Layer (Type) | Output Shape | Params # |
|--------------|--------------|----------|
| bidirectional_1 | (bs,1,90) | 68400 |
| bidrectional_2 | (bs,21) | 9408 |
| dense_1 (Dense) | (bs,128) | 2816 |
| dense_2 (Dense) | (bs,526) | 67854 |
| dense_3 (Dense) | (bs,256) | 134912 |
| dense_4 (Dense) | (bs,1) | 257 |

**Table 3. Architecture of LSTM based model**

Unidirectional LSTM processes data based on the preserved information only from the past. In problems where all timesteps of the input sequence are available, Bidirectional LSTMs train two instead of one LSTMs on the input sequence. The first on the input sequence as-is and the second on a reversed copy of the input sequence. This can provide additional context to the network and result in faster and even fuller learning on the problem.[4]

#### B. 1D CNN

1D Convolutional Neural Networks are also useful for extracting temporal features. The usage of 1D CNN has been recently explored for time series analysis[5]. 1D convolutional neural networks can be used for extracting local 1D patches (subsequences) from sequences and can identify local patterns within the window of convolution. Since the same transformation is applied on every patch identified by the window, a pattern learnt at one position can also be recognized at a different position, making 1D conv nets translation invariant[9]. Another use is to combine 1D conv nets with RNNs. If we have long sequence to process such that it cannot be realistically processed by RNNs, 1D conv nets can be used as pre-processing step to make the sequence smaller through downsampling by extracting higher level features, which can, then be passed on to the RNN as input. However, in our case, 1D conv nets with one extra dense net for classification works satisfactorily.

| Layer (type) | Output Shape | Param # |
|---|---|---|
| conv1d_1 (Conv1D) | (bs,137,32) | 288 |
| conv1d_2 (Conv1D) | (bs,133,32) | 5152 |
| max_pooling1d_1 (MaxPooling1D) | (bs,44,32) | 0 |
| conv1d_3 (Conv1D) | (bs,44,,32) | 6208 |
| conv1d_4 (Conv1D) | (bs,42,64) | 12352 |
| Conv1d_5 (Conv1D) | (bs,38,64) | 12352 |
| max_pooling1d_2 (MaxPooling1D) | (bs,12,64) | 0 |
| conv1d_6 (Conv1D) | (bs,11,128) | 16512 |
| conv1d_7 (Conv1D) | (bs,10,128) | 32896 |
| conv1d_8 (Conv1D) | (bs,9,256) | 65792 |
| global_average_pooling1d_1 | (bs,256) | 0 |
| Dropout_1 (Dropout) | (bs,256) | 0 |
| Dense_1 (Dense) | (bs,1) | 256 |

**Table 4. Architecture of CNN based model**

IV. EXPERIMENTS

All of our architectures are trained and tested on PC which has i7-7th generation intel processor with 16GB RAM and this PC also has Nvidia GTX 1070Ti.

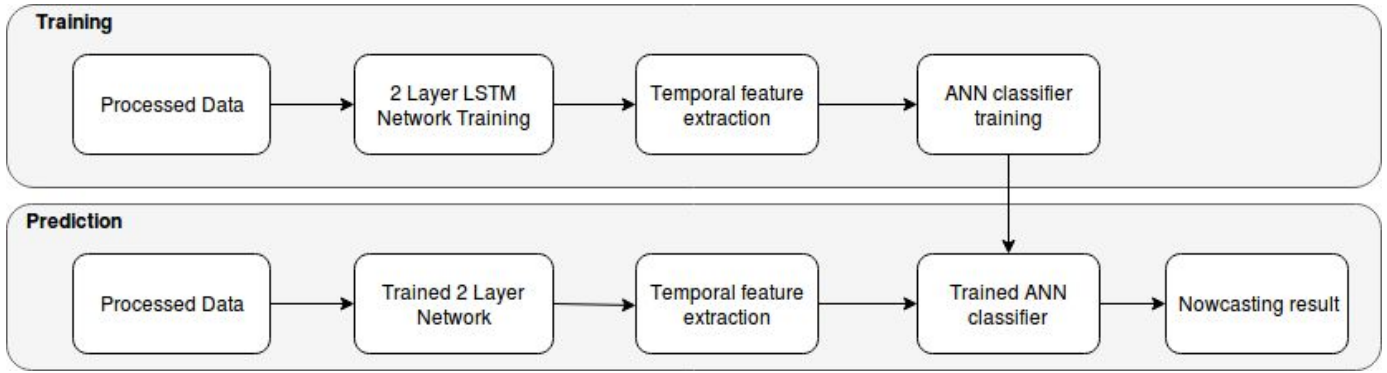

**Figure B. Flow chart of LSTM based working model**

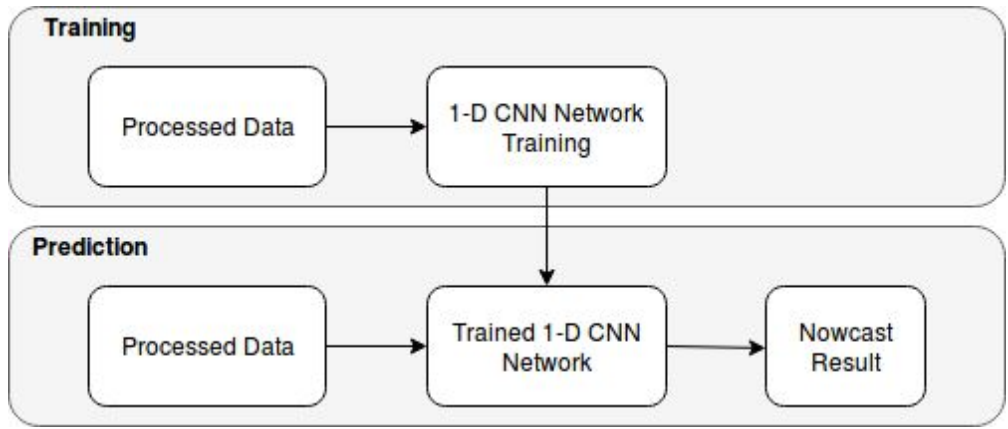

**Figure C. Flow chart of CNN based working model**

The details of the our proposed architecture is given in Table 3 where bs denotes the BatchSize. Here we have used a cascade of two BiLSTMs. The output of the BiLSTM network is given as an input to a second bidirectional LSTM with reduced number of parameters. After training both the bidirectional LSTMs, we extract the temporal features from the last BiLSTM which are further used to train a dense layer multi perceptron network (also known as fully connected layers) as a classifier.

The activation function used after each fully connected neural network is ReLu[6][8]. The total trainable parameters in this model are 283,647. Hence it is easy and faster to train on a larger dataset.

The details of the CNN based architecture is shown in Table 4. Apart from Conv-1D, we introduced two max pooling layers after 2nd and 5th convolutional layer because it contributed a lot of redundancy in input data. A global average pooling layer was added at the end of the 8th convolutional layer to connect it with the dense layer. While performing experiments, we noticed that this model often led to overfitting for the given dataset. This was reduced by adding 40% dropout[12] just before last fully connected layer. The total trainable parameters are 151,809, which are almost half of that used in our first architecture. However, it takes more time for the model to converge for attaining an accuracy near to that of the first model. This type of behaviour could be attributed to the addition of dropout in the last layer.

In Table 5, we have shown experiments run on two different models. We observe that the LSTM based model achieved an accuracy of ~93.7% for predicting precipitation for next hour on Indian weather dataset and ~93.2% accuracy for the next to next hour. Using second architecture, we get a difference of around 4% in the accuracy. Also, for the dataset available on Kaggle, we achieved an accuracy of ~89.8 on LSTM based model and ~84.9% on CNN based model for one time step.

In the figures from D to I, we have shown different loss and accuracy graphs for two different architectures on two datasets. We can say that LSTM based approach is more stable compared to 1D CNN based approach.

| North America weather dataset (selected cities) | Input hours=24, output hours=1 | Input hours=24, output hours=2 | Input hours=12, output hours=1 | Input hours=12, output hours=2 |
|---|---|---|---|---|
| Accuracy on LSTM | 88.97% | 87.33% | 88.86% | 87.19% |
| Accuracy on CNN | 84.97% | 84.68% | 84.23% | 82.88% |
| Indian weather dataset | Input hours=24, output hours=1 | Input hours=24, output hours=2 | Input hours=12, output hours=1 | Input hours=12, output hours=2 |
| Accuracy on LSTM | 93.7% | 93.26% | 91.24% | 90.54% |
| Accuracy on CNN | 89.83% | 89.81% | 88.96% | 88.12% |

**Table 5. LSTM and CNN based model analysis**

## V. RESULTS

Architectures developed using CNN has almost half of the trainable parameters compared to the architecture which is developed using LSTM but at the cost of some accuracy. 1D CNN takes a longer time to converge to attain accuracy near to that of a bidirectional LSTM. It has also been observed that the loss function converges smoothly in the case of LSTM ( Figure G) while the curve for 1D CNN has frequent spikes before it converges to a minimum (Figure F).

## VI. CONCLUSION

In this paper, we have explored two deep learning techniques for precipitation nowcasting that can effectively bridge the gap between Numerical Weather Prediction and basic extrapolation techniques.

LSTM and 1D CNN models have both proved to be very effective in weather prediction problem as both of them preserve the temporal features in our time series dataset. However, the LSTM model has given more promising results than 1D CNN.

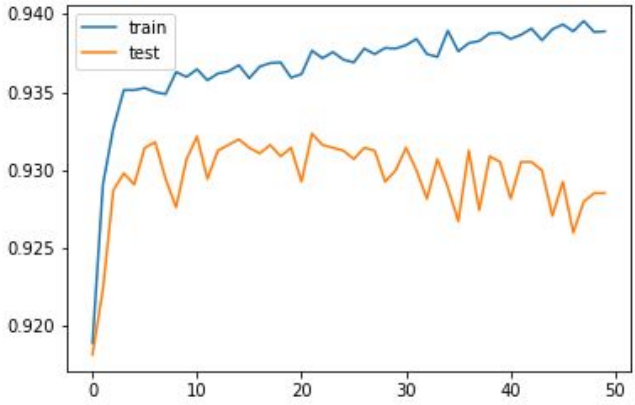

**Figure D.** Accuracy graph on Indian weather patterns (LSTM based architecture)

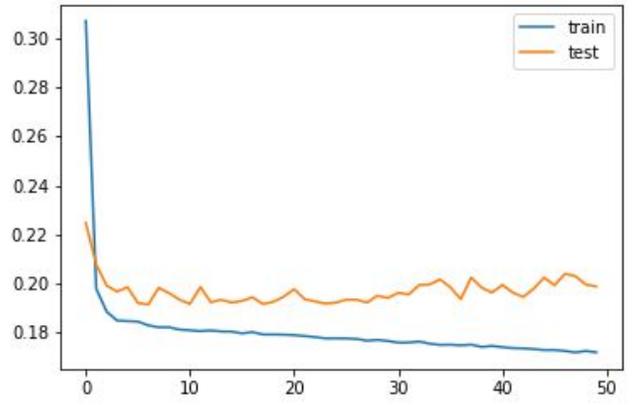

**Figure E.** Loss graph on Indian weather patterns (LSTM based architecture)

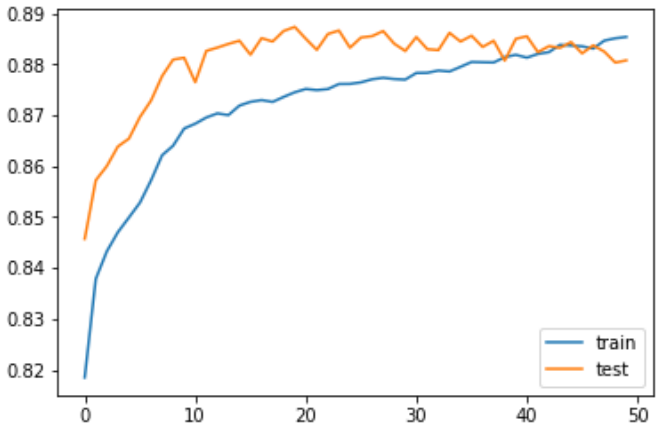

**Figure F.** Accuracy graph on Kaggle dataset (LSTM based architecture)

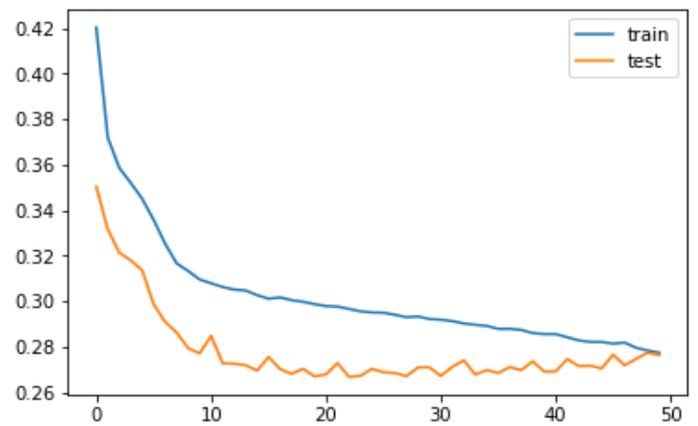

**Figure G.** Loss graph on Kaggle dataset (LSTM based architecture)

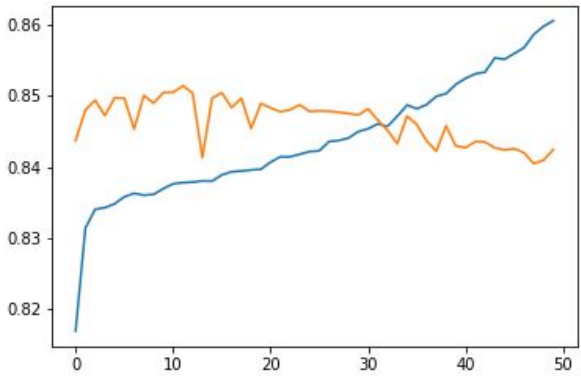

**Figure H.** Accuracy graph on Kaggle dataset (CNN based architecture)

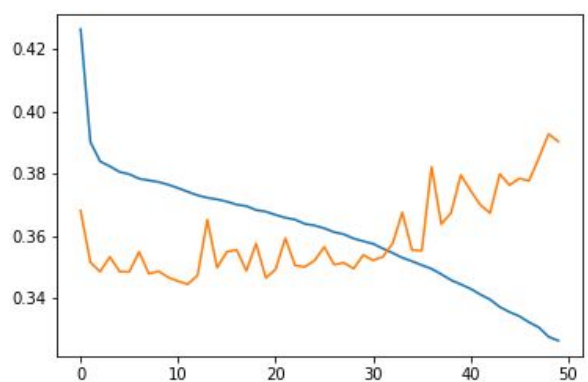

**Figure I.** Loss graph on Kaggle dataset (CNN based architecture)